\documentclass[10pt,twocolumn,letterpaper]{article}

\usepackage{cvpr}              

\definecolor{cvprblue}{rgb}{0.21,0.49,0.74}
\usepackage[pagebackref,breaklinks,colorlinks,allcolors=cvprblue]{hyperref}
\usepackage{makecell}
\usepackage{pifont}
\usepackage{multirow}

\def\paperID{14537} 
\def\confName{CVPR}
\def\confYear{2025}

\title{Mitigating Ambiguities in 3D Classification with Gaussian Splatting\thanks{The work was supported in part by the National Key Research and Development Project of China (2022YFF0902402) and NSFC (Grant 62473193).}}

\author{
Ruiqi Zhang$^{1,\dagger}$, Hao Zhu$^{1,\dagger}$, Jingyi Zhao$^2$, Qi Zhang$^3$, Xun Cao$^{1,*}$, Zhan Ma$^{1,*}$\\
$^1$ Nanjing University, $^2$ Imperial College London, $^3$ Vivo Company\\
$\dagger$ Equal contribution. * Corresponding author: {\tt \{caoxun, mazhan\}@nju.edu.cn}
}

\begin{document}
\maketitle
\begin{abstract}
3D classification with point cloud input is a fundamental problem in 3D vision. However, due to the discrete nature and the insufficient material description of point cloud representations, there are ambiguities in distinguishing wire-like and flat surfaces, as well as transparent or reflective objects. To address these issues, we propose Gaussian Splatting (GS) point cloud-based 3D classification. We find that the scale and rotation coefficients in the GS point cloud help characterize surface types. Specifically, wire-like surfaces consist of multiple slender Gaussian ellipsoids, while flat surfaces are composed of a few flat Gaussian ellipsoids. Additionally, the opacity in the GS point cloud represents the transparency characteristics of objects. As a result, ambiguities in point cloud-based 3D classification can be mitigated utilizing GS point cloud as input. To verify the effectiveness of GS point cloud input, we construct the first real-world GS point cloud dataset in the community, which includes 20 categories with 200 objects in each category. Experiments not only validate the superiority of GS point cloud input, especially in distinguishing ambiguous objects, but also demonstrate the generalization ability across different classification methods. Our project page: \href{https://ruiqi-nju.github.io/MACGS}{https://ruiqi-nju.github.io/MACGS}.
\end{abstract}  

\begin{figure}[t]
  \centering
   \includegraphics[width=0.95\linewidth]{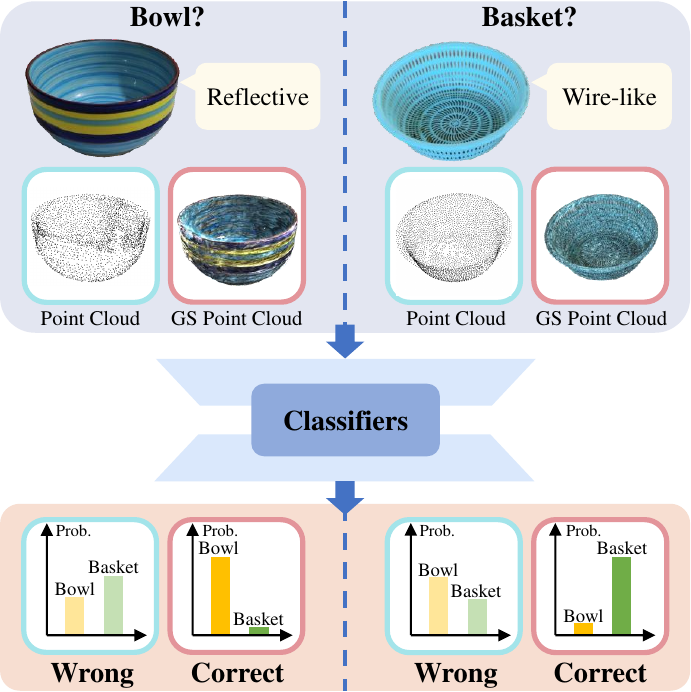}
   \caption{Illustration of ambiguity mitigation in 3D classification using GS point cloud. Although the Bowl and the Basket have different surface types (flat \textit{vs} wire-like, specular \textit{vs} diffuse reflection), existing classifiers often confuse them due to similar shapes when described as traditional point cloud, resulting in wrong probability predictions. These ambiguities can be mitigated by introducing GS point cloud.
   }
   \label{fig:first}
   \vspace{-0.3cm}
\end{figure}

\section{Introduction}
\label{sec:introduction}

\textit{``The performance of machine learning methods is heavily dependent on the choice of data representation (or features) on which they are applied.''} 
\hspace{2cm}Bengio \etal~\cite{bengio2013representation}

3D classification, which aims at identifying and categorizing objects with 3D point cloud input, is gaining increasing attention due to the convenience of 3D acquisition equipment and its better ability to characterize structure and handle occlusions compared to traditional 2D images. Different from the regular grids in 2D images, the 3D point cloud is organized in an irregular, unordered way, leading to the development of various neural network architectures~\cite{qi2017pointnet,qi2017pointnet++,qian2022pointnext,lan2019modeling,hassani2019unsupervised,lei2019octree,wang2023octformer,wu2024point} for extracting local shapes.

However, due to the discrete nature and hard description of existence in point cloud representation, there are two inherent ambiguities in characterizing local shapes and appearance in existing point cloud-based classification methods. For the first ambiguity, the sampling rate of existing 3D acquisitions is often insufficient for characterizing high-frequency structures, resulting in similar point cloud representations for wire-like and flat surfaces of objects that belong to different categories. For the second ambiguity, existing point cloud representation assumes that the components of a 3D object follow a hard existing (1) or non-existing (0) constraint, without any intermediate state, which violates the description of transparent and reflective objects in the real world. As a result, existing point cloud-based classification methods may struggle to distinguish between objects with wire-like and flat surfaces, as well as those with similar structures but different transparency and reflectivity.

To mitigate the ambiguities in point cloud-based 3D classification, we propose Gaussian Splatting (GS)~\cite{kerbl3Dgaussians} point cloud-based classification. GS point cloud is originally designed for rendering realistic images of novel views with a set of images captured from different positions. To achieve this goal, particularly in compensating for holes caused by occlusion and modeling transparent or reflective objects, each 3D point is represented as a continuous 3D Gaussian kernel function, with an additional opacity coefficient to characterize each point's soft existence. We note that the incorporation of these GS coefficients compensates for the coarse shape representation in point cloud, thereby enhancing the accuracy of classification models. \cref{fig:first} illustrates that integrating GS point cloud effectively mitigates ambiguities in 3D classification.

To verify the efficacy of GS point cloud in 3D classification, we build the first large-scale real-world dataset in the community based on an existing multi-view image dataset~\cite{yu2023mvimgnet}. The proposed dataset contains 4,000 objects from 20 categories, including easily misclassified objects such as the wire-like, flat, and reflective objects mentioned above. On this basis, several classical and state-of-the-art classification methods are used for comparison, where only the network input is replaced from pure point cloud (\textit{i.e.}, position) to GS point cloud (\textit{i.e.}, position, standard deviations, rotation matrix, and opacity), while other network components remain unchanged. We find that the introduction of Gaussian coefficients significantly improves classification for ambiguous objects such as Loudspeaker and Paper box (with similar cuboid structures but different wire-like and flat surfaces, respectively), and Mug and Ashcan (with similar structures but different reflectivity). Furthermore, by visualizing the global features in these neural networks using the t-SNE technique (see~\cref{fig:tsne}), we observe that the separability between different categories is significantly increased, which verifies the effectiveness of the added Gaussian coefficients in GS point cloud for 3D classification.
Specifically, the main contributions of the work include,

\begin{enumerate}
    \item As far as we know, this is the first exploration to enhance 3D task (namely classification) using GS point cloud. We provide a deep analysis of the ambiguities in traditional point cloud-based classification and how GS point cloud mitigates them theoretically.
    \item We construct the first large-scale real-world GS point cloud dataset, which provides the data base for various GS point cloud-based 3D tasks in future.
    \item Quantitative and qualitative experimental results demonstrate that the GS point cloud effectively mitigates ambiguities and improves classification accuracy across a wide range of methods.
\end{enumerate}

\section{Related Work}
\label{sec:related work}

\subsection{3D Classification}

3D classification with point cloud input has gained significant attention due to the widespread application, advancements in 3D acquisition equipment, and rapid development of deep learning techniques. Four main types of methods have been proposed to process the irregular and unordered structure of point cloud, \textit{i.e.}, the point-wise MLP, convolution-based, graph-based, and hierarchical data structure-based methods. The first method~\cite{qi2017pointnet,qi2017pointnet++,qian2022pointnext,ma2022rethinking} applies multiple shared Multi-Layer Perceptrons (MLPs) to extract features from the input point cloud matrix, then these features are aggregated. The second method~\cite{liu2019relation,liu2019densepoint,thomas2019kpconv,lan2019modeling} designs specific convolution operators to connect neighboring points in 3D space. By iteratively applying these operators, deep features are obtained for further analysis. The third method~\cite{wang2019dynamic,hassani2019unsupervised,shen2018mining} treats point cloud as a graph, where each point is a graph's vertex and direct edges are built between neighboring points, then a graph neural network is applied. The fourth method~\cite{lei2019octree,klokov2017escape,li2018so} organizes the point cloud into hierarchical structures, where deep features are extracted from leaf nodes to the root node hierarchically. 

\textit{Although numerous methods have been proposed, the inherent ambiguities in point cloud representation for classifying wire-like, flat surfaces, and transparent or reflective objects remain unaddressed.}

\subsection{3D Gaussian Splatting}

GS~\cite{kerbl3Dgaussians} is an emerging technique for novel view synthesis build on point cloud-based rendering. To overcome the limitations of pure point cloud for representing view-dependent textures of 3D objects, opacity and spherical harmonics~\cite{yu2021plenoctrees} are attached to each point's attributes. However, due to the discrete nature of point cloud, there are many holes in the rendered image. To address this issue, each point is modeled using a 3D anisotropic Gaussian function. This allows the textures of empty spaces to be represented, effectively mitigating the holes problem. 

Due to the high-quality reconstruction and high-efficiency rendering, GS dominates the area of novel view synthesis and has been further optimized by subsequent research. Mip-Splatting~\cite{Yu2024MipSplatting} constrains the frequency content of 3D representations to below half the maximum sampling frequency, effectively mitigating high-frequency artifacts resembling Gaussian shapes. Scaffold-GS~\cite{scaffoldgs} initializes a voxel grid with learnable features assigned to each voxel point, where the attributes of the Gaussians are derived from interpolated features and lightweight neural networks, resulting in more accurate view-dependent effects while reducing the model's storage. HAC~\cite{hac2024} builds upon the concept of Scaffold-GS~\cite{scaffoldgs} by representing the scene using anchor points, each associated with learnable features. Additionally, it introduces an adaptive quantization module that compresses the anchor point features through a multi-resolution hash grid~\cite{muller2022instant}.

\textit{However, to the best of the authors knowledge, existing works mainly focus on improving the rendering quality or efficiency of GS, and none of them think deeply on how this new representation can empower traditional computer vision tasks. Compared with these works, we build the first large-scale real-world GS dataset for classification and provide a deep analysis both qualitatively and quantitatively.}

\section{GS Point Cloud-based 3D Classification}
\label{sec:method}

In this section, we first introduce the background of GS point cloud representation. Next, we describe the pipeline used throughout the paper of 3D classification with both point cloud and GS point cloud inputs. Finally, we analyze the advantages of GS point cloud-based classification for mitigating ambiguities.
\subsection{GS Point Cloud}
Given a 3D object, traditional point cloud represents it using multiple isolated points and records each point's position $\boldsymbol{p}=[x,y,z]^{\top}$ and color $\boldsymbol{c}=[r,g,b]^{\top}$. GS point cloud differs from traditional one by treating the object as a combination of multiple 3D Gaussian distributions. As a result, more information for characterizing the Gaussian distribution is recorded, namely, the standard deviation $\boldsymbol{s}=[s_x,s_y,s_z]^{\top}$ along three axes and the rotations around the axes. Note that, because the standard deviation is also called scale in GS-related papers, these two terms are used interchangeably in the following sections. The quaternion $\boldsymbol{q}=[q_1,q_2,q_3,q_4]^{\top}$ is widely used here to represent rotation, as it avoids ambiguities associated with Eular angle representation. Additionally, GS point cloud softens the opacity $\boldsymbol{o}\in [0,1]$ to describe the material features of each point, unlike the constant $\boldsymbol{o}=1$ in traditional point cloud. Finally, GS point cloud extends the pure color to a color function, \textit{i.e.}, a series of spherical harmonics, for characterizing view-dependent colors.

\subsection{Pipeline of 3D Classification}
\label{subsec:pipeline}
A point cloud can be represented as $(\mathbf{P}\triangleq\{\boldsymbol{p}_i\}_{i=1}^{N},\boldsymbol{y})$, where $\boldsymbol{p}_i=[x,y,z]^{\top}$ represents the coordinate of the $i$-th point and $N$ is the total number of points in the cloud. The ground-truth label $\boldsymbol{y}\in\{1,2,\ldots,k\}$ indicates one of $k$ possible classes. Given such a point cloud, classical learning-based classification methods $\mathbf{F}_{\boldsymbol{\theta}}(\cdot)$ take the coordinates $\{p_i\}_{i=1}^{N}$ as input. The output is the probability distribution over $k$ categories,
\begin{equation}
    \mathbf{F}_{\boldsymbol{\theta}}(\mathbf{P})\triangleq\{F_{\boldsymbol{\theta},j}\}_{j=1}^{k},
    \label{eqn:cls_network}
\end{equation}
where $F_{\boldsymbol{\theta},j}$ is the probability belonging to the $j$-th class. To supervise the training of parameters $\boldsymbol{\theta}$ in the classification model, the cross-entropy loss function between $\mathbf{F}_{\boldsymbol{\theta}}(\mathbf{P})$ and $\boldsymbol{y}$ is often used.

\begin{figure}[!t]
\centering
\includegraphics[width=\linewidth]{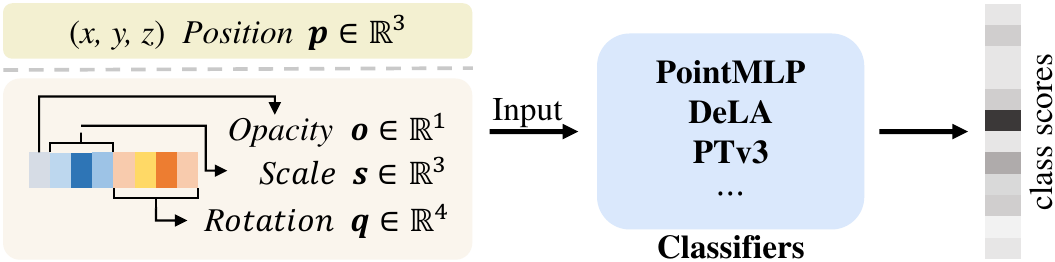} 
\caption{Pipeline of 3D classification used throughout the paper.}
\label{fig:pipeline}
\vspace{-0.3cm}
\end{figure}

For GS point cloud-based classification, we adopt the traditional point cloud-based classification pipeline, modifying only the input to include GS coefficients. Although GS point cloud has multiple coefficients, we use only four here, \textit{i.e.}, position, standard deviation, rotation, and opacity, namely, $(\mathbf{P}_{GS}\triangleq \{\boldsymbol{p}_i,\boldsymbol{o}_i,\boldsymbol{s}_i,\boldsymbol{q}_i\}_{i=1}^{N},\boldsymbol{y})$. Note that spherical harmonics are excluded to ensure a fair comparison, as traditional point cloud do not include color attributes. \cref{fig:pipeline} visualizes the pipeline of 3D classification used throughout the paper.

\subsection{Mitigating Ambiguities in 3D Classification} 
\label{sec:amb_theory_analysis}
Considering the discrete nature and hard description of existence in point cloud representation, there are two main ambiguities in characterizing local shapes and appearances in point cloud-based classification methods. The first ambiguity arises from the insufficient sampling rate of point cloud, making it difficult to accurately capture local geometric structures and surface details of objects. This can lead to inconsistencies in the classification results when dealing with objects that have similar global shapes but different local features, such as the wire-like and flat surfaces. The second ambiguity in appearance-related characterization is caused by the hard existing constraint in point cloud, resulting in representation that struggles to differentiate objects with similar structures but varying transparency and reflectivity. These ambiguities limit the performance of existing classifiers in distinguishing objects with similar structures but different local shapes and material properties.

\begin{figure}[ht]
  \centering
   \includegraphics[width=\linewidth]{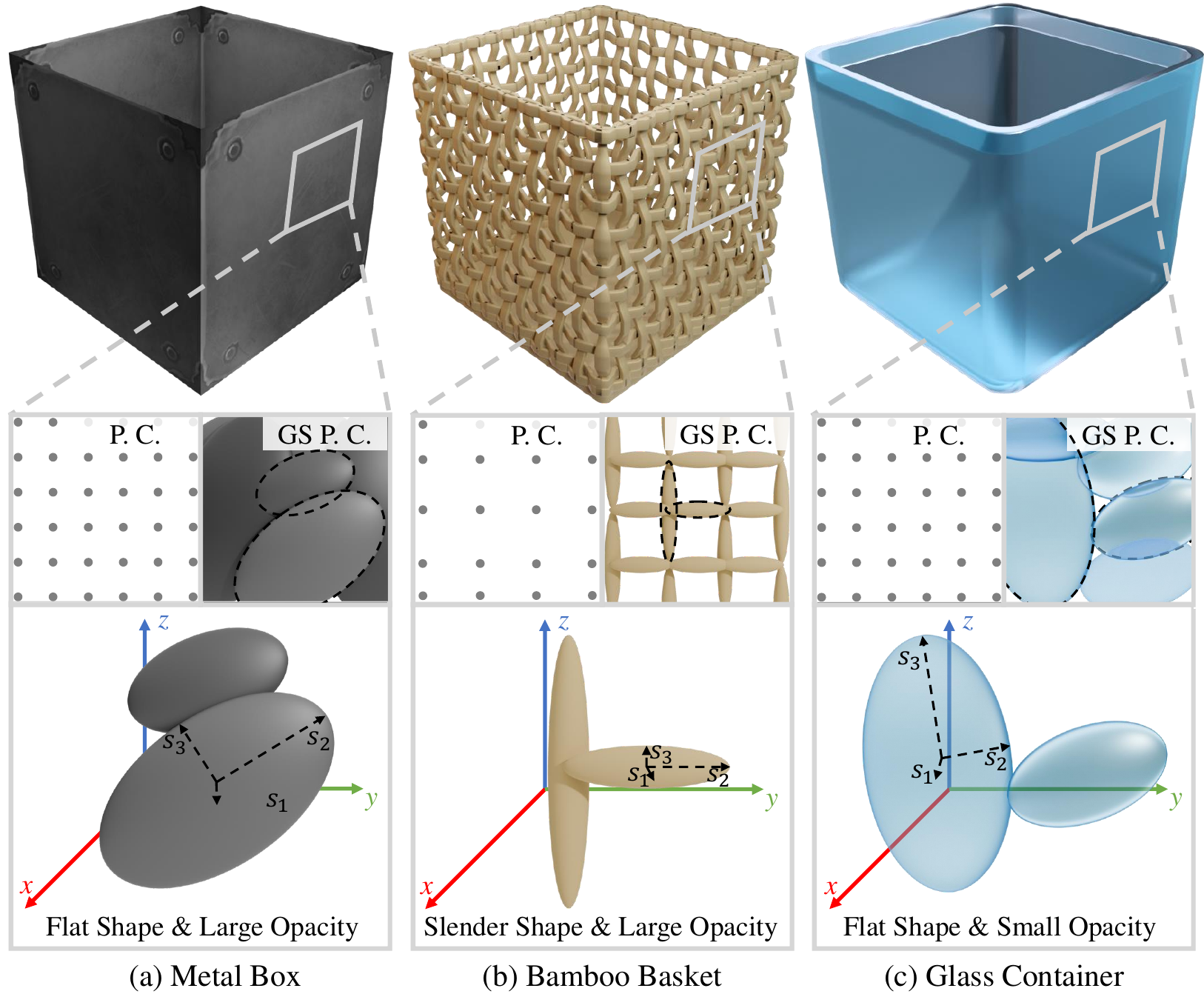}
   \caption{Comparison between point cloud and the ellipsoid representation in Gaussian Splatting (GS). These three examples illustrate how opacity, scale, and rotation of Gaussian coefficients represent object transparency and surface structure. In the left example, a series of large, flat ellipsoids with high opacity represent the flat, opaque surface of the metal box. In the middle example, adjustments in scale and rotation enable ellipsoids to follow the intricate, wire-like structure of the bamboo basket. The right example uses low-opacity, flat ellipsoids to represent the smooth, transparent surface of the glass container.}
   \label{fig:principle}
   \vspace{-0.3cm}
\end{figure}

According to the definition mentioned in~\cref{subsec:pipeline}, the GS-based point cloud representation can mitigate the ambiguities of traditional point cloud-based classification in two aspects,
\begin{enumerate}
    \item \textbf{Ambiguity in local shape characterization}. Due to the discrete nature of traditional point cloud, there is an ambiguity in distinguishing the wire-like surfaces from flat ones due to the lack of information about the local geometry. In contrast, GS-based point cloud offers a continuous representation, where each GS point not only describes its own position but also encodes information about its neighboring space. By representing the point cloud as Gaussian ellipsoids, where the covariance matrix defines the scale and rotation of each ellipsoid, the GS-based classification provides a more robust and accurate description of local shapes, effectively addressing the first ambiguity. As shown in~\cref{fig:principle}, the scale controls the size of the ellipsoid, while the rotation determines its orientation, and together they describe the local geometry of the surface. For example, the wire-like surface in ~\cref{fig:principle}(b) is often represented by multiple slender ellipsoids, while flat surfaces like~\cref{fig:principle}(a) are represented by a small number of broader, flat ellipsoids. Consequently, the GS-based framework classifies them by constructing a continuous representation of the point cloud, which captures both the global and local geometric features of objects, allowing for a more discriminative classification.
    
    \item \textbf{Ambiguity in appearance-related characterization}. In traditional point cloud representation, each coordinate refers to an opaque point in 3D space, resulting in ambiguity when distinguishing objects with similar shapes but different transparency. GS point cloud softens this constraint by relaxing the constant opacity of 1 to a variable opacity in the range $[0,1]$. Comparing ~\cref{fig:principle}(a) and ~\cref{fig:principle}(c) reveals that although the metal box and glass container share similar flat surfaces and exhibit comparable point cloud distributions, the GS point cloud effectively differentiates these surface types by representing scene transparency through opacity. This capability enhances 3D classification for transparent or reflective objects, providing greater clarity and consistency in point cloud data, and enabling more accurate classification. 

\end{enumerate}

\section{Experiments}
\label{sec:experiment}

The previous section has explained the effectiveness of additional coefficients of GS point cloud for 3D classification in theory. In this section, we focus on validating these analyses through experiments.

\subsection{Experiments Setup}

\begin{table*}[t]
\small
\centering
\begin{tabular}{lcccccc}
\toprule
 {Dataset} & {\# of objects} & {\# of categories} & {Real} & \makecell[c]{Local Structure\\Characterization} & \makecell[c]{Transparency\\Characterization} & \makecell[c]{View-dependent\\Color} \\
\midrule
  {ShapeNet~\cite{shapenet2015}} & {51k} & {55} & {\ding{55}} & {\ding{55}} & {\ding{55}} & {\ding{55}}\\
  {ModelNet~\cite{Zhirong15CVPR}} & {12k} & {40} & {\ding{55}} & {\ding{55}} & {\ding{55}} & {\ding{55}}\\
  {ScanObjectNN~\cite{uy-scanobjectnn-iccv19}} & {14k} & {15} & {\ding{51}} & {\ding{55}} & {\ding{55}} & {\ding{55}}\\
  {MVPNet~\cite{yu2023mvimgnet}} & {87k} & {150} & {\ding{51}} & {\ding{55}} & {\ding{55}} & {\ding{55}}\\
  {GS dataset (ours)} & {4k} & {20} & {\ding{51}} & {\ding{51}} & {\ding{51}} & {\ding{51}}\\
\bottomrule
\end{tabular}
\caption{Comparison of other point cloud datasets with our proposed GS point cloud dataset.}
\label{table_dataset}
\vspace{-0.3cm}
\end{table*}

\noindent \textbf{Dataset.} Existing point cloud datasets~\cite{shapenet2015, Zhirong15CVPR, uy-scanobjectnn-iccv19, yu2023mvimgnet} provide only 3D positions and lack coefficients related to GS, making them unsuitable for our task. To validate the effectiveness of GS point cloud, we build the first large-scale real-world GS point cloud dataset, based on the MVImageNet dataset~\cite{yu2023mvimgnet}. MVImageNet consists of 219,188 objects from 238 classes, where each object is represented by multi-view images and their corresponding binary masks. Using the official GS implementation~\cite{kerbl3Dgaussians}, we generate GS point cloud from multi-view images. However, view-specific masks frequently introduce inconsistencies between views, as they are generated independently by a third-party segmentation method~\cite{carvekit}. These inconsistencies degrade the quality of the reconstructed GS point cloud, complicating 3D classification evaluation. To address these challenges, we manually curated a subset of 4,000 objects, selecting 20 categories from the initial set of 238, with 200 examples per category. Each object was associated with approximately 30 masks, a labor-intensive process requiring careful attention to ensure consistency across views. We structured the data meticulously and performed object-by-object training, which took over 600 GPU hours on a single A100 card.

Our GS point cloud dataset stands out from traditional point cloud datasets by providing local structure characterization that captures finer geometric details in the scene through scale and rotation embedded in each point, as well as transparency characterization that conveys information about an object's transparency and reflectivity. While our research focuses on point-based classification and does not incorporate texture-related spherical harmonics, our dataset retains these coefficients. Compared to traditional point cloud rendering, our GS point cloud dataset enables high-fidelity, view-dependent image rendering, which significantly facilitates multiview-based classification methods~\cite{su2015multi,yang2019learning,feng2018gvcnn,wei2020view,hamdi2021mvtn}, enhancing the classification of real-world objects with diverse material properties.

\noindent \textbf{Baseline Algorithms.} 
In our experiments, we select a set of representative point-based methods for validation. These include the groundbreaking PointNet~\cite{qi2017pointnet} and PointNet++~\cite{qi2017pointnet++}, which process point cloud using point-wise MLP. Additionally, we compare state-of-the-art methods developed upon PointNet/PointNet++, such as PointNeXt~\cite{qian2022pointnext}, which enhances performance through an optimized training strategy, and PointMLP~\cite{ma2022rethinking}, which significantly boosts inference speed with a lightweight local geometry extractor. Finally, we also consider state-of-the-art methods from other categories, including convolution-based approaches like DeLA~\cite{chen2023decoupled} and transformer-based methods such as PTv3~\cite{wu2024point}, both of which have achieved impressive results in their respective domains.

In implementation, all results are obtained by re-training the official codes and following official guidelines released by authors with a 24G 3090 GPU. 
All of these methods are combined with both traditional point cloud and GS point cloud inputs. When verifying the effectiveness of GS point cloud input, we modifies only the first layer of the network to accommodate the increased channels, \textit{i.e.}, from 3 (only position) to 4 (position+opacity), 10 (position+scale+rotation), 11 (position+opacity+scale+rotation). 

\begin{table}[t]
\setlength\tabcolsep{3pt}
\small
\centering
\begin{tabular}{l|cc|cc}
\toprule
\multirow{2}{*}{Method} & \multicolumn{2}{c|}{w/o. GS coeffs.} & \multicolumn{2}{c}{w. GS coeffs.}\\
& OA (\%) & mAcc (\%) & OA (\%) & mAcc (\%)\\
\midrule
  {PointNet} & {73.56} & {73.77} & {80.87 (\textbf{+7.31})} & {81.46 (\textbf{+7.69})} \\
  {PointNet++} & {83.02} & {82.17} & {86.63 (\textbf{+3.61})} & {86.18 (\textbf{+4.01})} \\
  {PointNeXt} & {87.77} & {86.54} & {89.78 (\textbf{+2.01})} & {88.70 (\textbf{+2.16})} \\
  {PointMLP} & {87.91} & {86.75} & {90.21 (\textbf{+2.30})} & {89.48 (\textbf{+2.73})} \\
  {DeLA} & {88.78} & {87.92} & {90.36 (\textbf{+1.58})} & {89.41 (\textbf{+1.49})} \\
  {PTv3} & {88.78} & {87.88} & {89.93 (\textbf{+1.15})} & {88.46 (\textbf{+0.58})} \\
\bottomrule
\end{tabular}
\caption{Comparison of overall accuracy and mean accuracy for different baseline methods before and after incorporating Gaussian coefficients.}
\label{table_classification}
\vspace{-0.3cm}
\end{table}

\subsection{Results}
\cref{table_classification} provides a comprehensive comparison of the overall accuracy (OA) and the mean class accuracy (mAcc) across different methods, both with and without GS coefficients. As observed, all methods show an improvement in both OA and mAcc when GS point cloud is employed. For earlier works such as PointNet, the integration of GS coefficients leads to a notable improvement of over 7\% in both metrics. In contrast, the state-of-the-art methods, including PointNeXt, PointMLP, DeLA, and PTv3, also demonstrate improvements, although to a lesser extent. The OA for these methods reaches approximately 90\%, with the minimum improvement being 1\% over the raw point cloud inputs.

\begin{figure*}[!t]
\centering
\includegraphics[width=\linewidth]{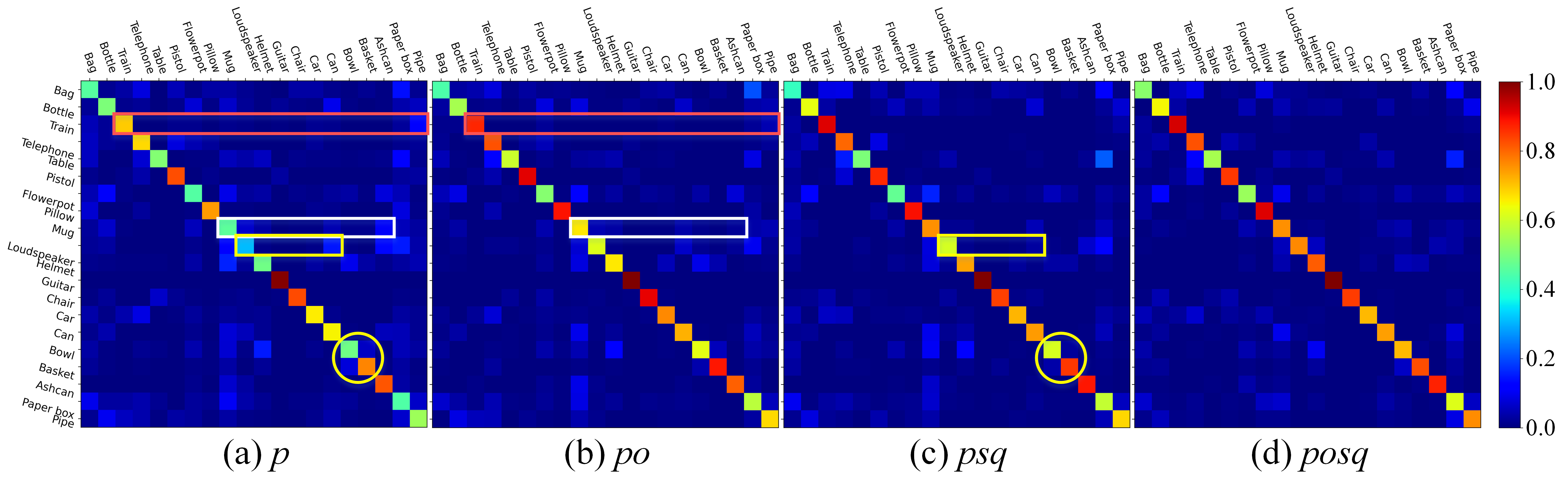} 
\caption{Comparisons of the probabilities output from the PointNet with different inputs. From left to right: only position, position+opacity, position+scale+rotation, position+opacity+scale+rotation. For each sub-figure, the color of the $i$-th row and $j$-th col represents the probability of the $i$-th object be classified as the $j$-th one. The rightmost image is the colorbar.}
\label{fig:prob_evo}
\end{figure*}

To further analyze the effectiveness of different coefficients in GS point cloud, we feed the combinations of position+opacity ($po$), position+scale+rotation ($psq$) and position+opacity+scale+rotation ($posq$) into existing networks, respectively. \cref{tab:softmax_probability} provides detailed classification probabilities of the correct class by PointNet for different categories, where the probabilities have increased to varying degrees. \cref{fig:prob_evo} shows the evolution of probability distribution with different inputs. A more detailed explanation of~\cref{tab:softmax_probability} and~\cref{fig:prob_evo} will be provided in the following sections.

\begin{table*}[t]
\small
\centering
\begin{tabular}{clcccccccccc}
\toprule
 & {GS inputs} & {Bag} & {Bottle} & {Train} & {Telephone} & {Table} & {Pistol} & {Flowerpot} & {Pillow} & {Mug} & {Loudspeaker}\\
\midrule
\multirow{4}*{\rotatebox{90}{Prob.}}
  & {$p$} & {0.45} & {0.50} & {0.69} & {0.67} & {0.51} & {0.82} & {0.45} & {0.74} & {0.46} & {0.31}\\
  & {$po$} & {0.44} & {0.55} & {0.86} & {0.82} & {\textbf{0.59}} & {\textbf{0.91}} & {0.51} & {0.88} & {0.66} & {0.62}\\
  & {$psq$} & {0.41} & {0.63} & {0.91} & {0.79} & {0.50} & {0.86} & {0.47} & {0.89} & {0.75} & {0.60}\\
  & {$posq$} & {\textbf{0.51}} & {\textbf{0.64}} & {\textbf{0.92}} & {\textbf{0.82}} & {0.55} & {0.84} & {\textbf{0.53}} & {\textbf{0.91}} & {\textbf{0.75}} & {\textbf{0.76}}\\
\bottomrule
\toprule
& {GS inputs} & {Helmet} & {Guitar} & {Chair} & {Car} & {Can} & {Bowl} & {Basket} & {Ashcan} & {Paper box} & {Pipe}\\
\midrule
\multirow{4}*{\rotatebox{90}{Prob.}}
  & {$p$} & {0.48} & {0.99} & {0.82} & {0.65} & {0.65} & {0.48} & {0.76} & {0.81} & {0.44} & {0.54}\\
  & {$po$} & {0.66} & {0.99} & {\textbf{0.90}} & {\textbf{0.76}} & {0.72} & {0.62} & {\textbf{0.88}} & {0.80} & {0.57} & {0.67}\\
  & {$psq$} & {0.73} & {0.99} & {0.84} & {0.71} & {0.74} & {0.61} & {0.85} & {\textbf{0.88}} & {0.58} & {0.68}\\
  & {$posq$} & {\textbf{0.80}} & {\textbf{0.99}} & {0.84} & {0.71} & {\textbf{0.74}} & {\textbf{0.71}} & {0.82} & {0.87} & {\textbf{0.61}} & {\textbf{0.75}}\\
\bottomrule
\end{tabular}
\caption{The average probability of the correct class obtained by PointNet for each category with different inputs.}
\label{tab:softmax_probability}
\vspace{-0.3cm}
\end{table*}

\subsection{Opacity}

\begin{figure*}[t]
\centering
\includegraphics[width=0.95\linewidth]{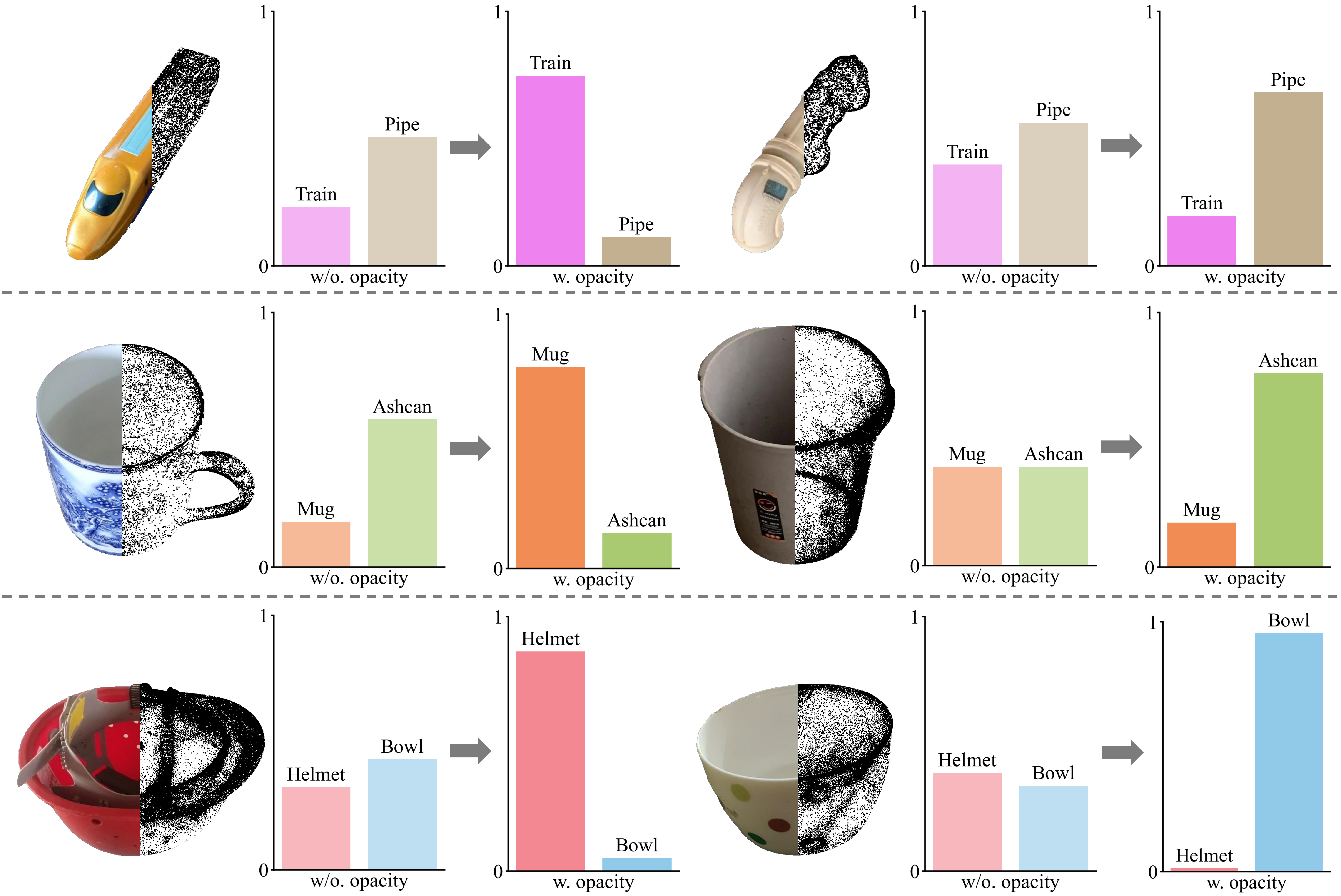} 
\caption{Comparison of the geometric similarity leading to misclassification between the helmet and the bowl (first row), and the classification probability map before and after using opacity, where red represents the correct class and blue represents other classes (second row).}
\label{fig:opacity}
\vspace{-0.3cm}
\end{figure*} 

\begin{figure*}[t]
\centering
\includegraphics[width=0.95\linewidth]{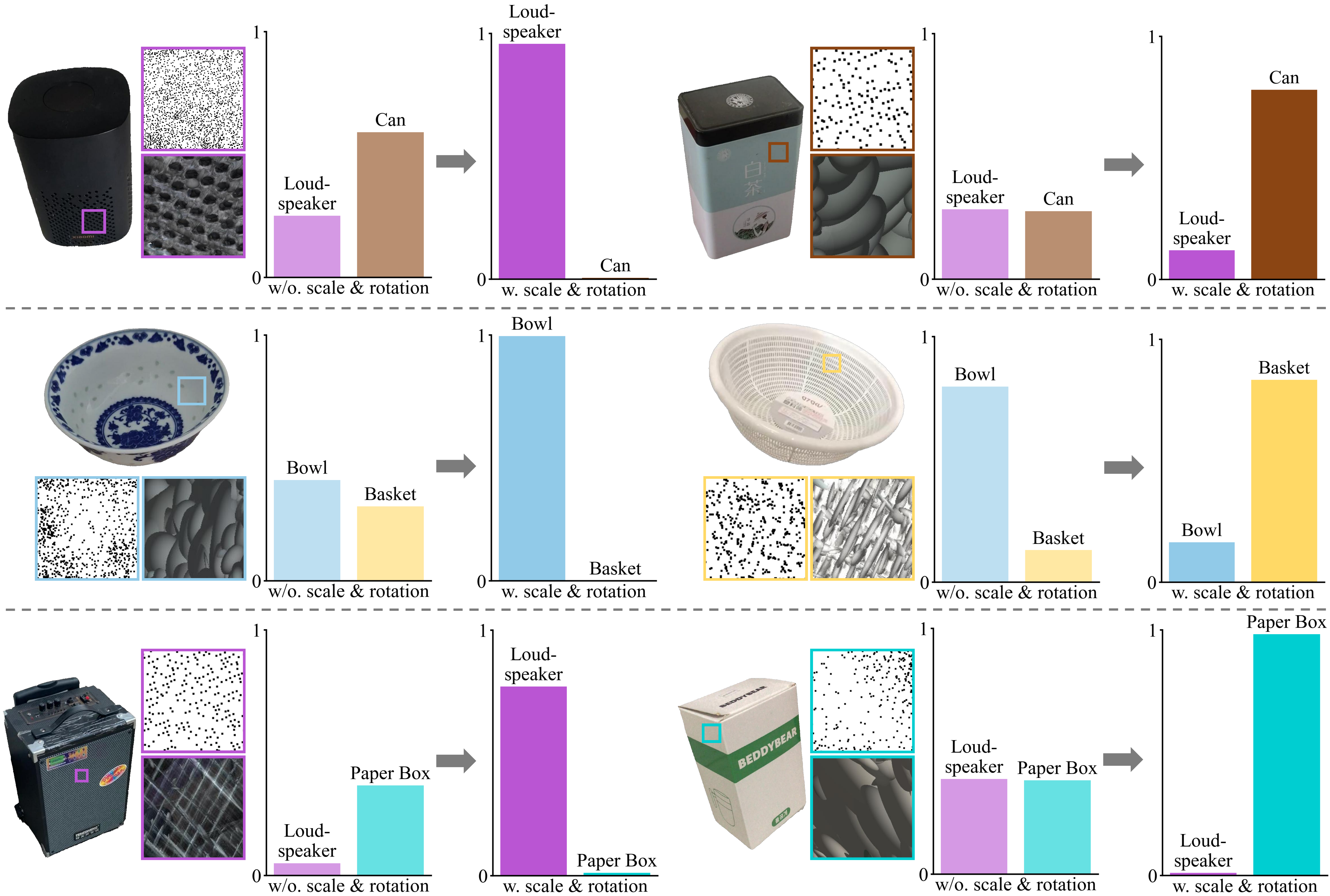} 
\caption{
Comparison of the probability of obtaining the correct category with or without using the scale and orientation of the GS point cloud. For each object, the right-top and right-bottom images visualize the point cloud and GS point cloud of the zoom-in area box. Two histograms right are the classification probabilities output from the PointNet without and with the scale and rotation coefficients in GS point cloud.}
\label{fig:size_orientation_ellipsoid}
\end{figure*}


According to~\cref{tab:softmax_probability}, the introduction of opacity significantly improves classification accuracy for objects with reflective or transparent properties, such as the Train, Mug, and Bowl. \cref{fig:prob_evo} visualizes these improvements. In the red rectangle of~\cref{fig:prob_evo}(a), the probability of correctly classifying the Train increases (\textit{i.e.}, from orange to red in the $3$-rd row and $3$-rd col block), while the probability of misclassification as the Pipe is also reduced (\textit{i.e.}, the $3$-rd row and $20$-th col block). The first row of~\cref{fig:opacity} visualizes the comparisons. It is observed that the shape of the Train is similar to the Pipe, \textit{i.e.}, both are strip-like objects. However, the Train is designed with a streamlined structure to reduce wind resistance, giving it a glossy surface with a lower opacity value. In contrast, the Pipe is typically made of plastic, which exhibits low reflectivity, resulting in an opacity value close to 1. As a result, the Train and the Pipe are classified more accurately when the additional opacity coefficient is added to the position.

\begin{figure*}[t]
  \centering
   \includegraphics[width=0.95\linewidth]{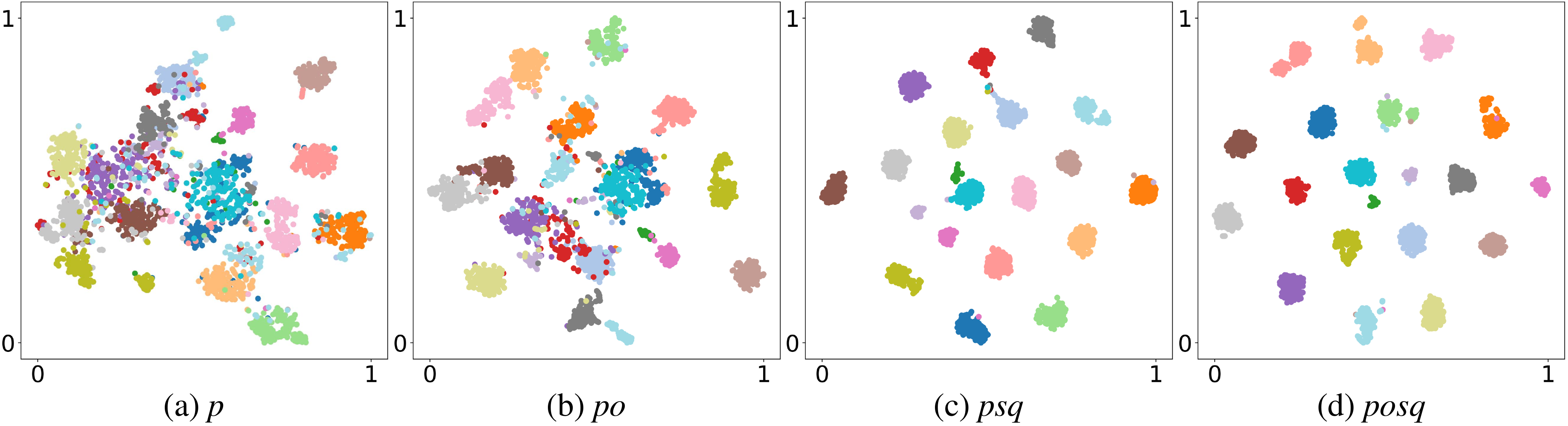}
   \caption{Visualizations of the global features in PointNet with different inputs using t-SNE technique. (a)-(d) are visualized features of network with only position, position+opacity, position+scale+rotation and position+opacity+scale+rotation inputs, respectively.}
   \label{fig:tsne}
   \vspace{-0.4cm}
\end{figure*}

Apart from the Train and the Pipe, it is also difficult to distinguish the Mug from the Ashcan if only the position is used. In~\cref{tab:softmax_probability}, when only position is used, the probability of correctly classifying the Mug is only $0.46$. One of the main reasons is the similar structure to the Ashcan. As demonstrated in the second row of~\cref{fig:opacity}, both the Mug and the Ashcan have a concave surface, which causes the model to have difficulty distinguishing them based on position alone. Although the Mug has a handle, PointNet assigns a higher probability to classifying it as the Ashcan rather than the correct category, Mug. Introducing opacity to the network significantly reduces this misclassification. As shown in the white rectangles in~\cref{fig:prob_evo}(a) and~\cref{fig:prob_evo}(b), the probability of correctly classifying the Mug improves significantly (\textit{i.e.}, from green to yellow in the $9$-th row and $9$-th col block), and the probability of misclassifying it as the Ashcan is also reduced (\textit{i.e.}, the $9$-th row and $18$-th col block). The change of the histogram in the second row of~\cref{fig:opacity} also verifies the effect of the opacity coefficient. 

\subsection{Scale and Rotation}

By comparing~\cref{fig:prob_evo}(a) and~\cref{fig:prob_evo}(c), it is observed that additional scale and rotation coefficients in GS point cloud help in classifying objects with wire-like surfaces as distinct from flat ones. For example, in the yellow rectangles, the probability of correctly classifying the Loudspeaker is significantly improved (\textit{i.e.}, from cyan to yellow in the $10$-th row and $10$-th col block), while the probability of misclassifying it as the Can is reduced (\textit{i.e.}, the $10$-th row and $15$-th col block). The first row of~\cref{fig:size_orientation_ellipsoid} visualizes the point cloud and GS representations of the Loudspeaker and the Can. Because the Loudspeaker shares a similar global structure to the Can, \textit{i.e.}, both are cylinder-like objects, it is difficult to distinguish them according to the point cloud (shown in the top-right boxes). As a result, the classifier, when fed with point cloud input, outputs unreliable results. 

However, the situation changes when introducing the Gaussian coefficients. As shown in~\cref{fig:size_orientation_ellipsoid}, since the Loudspeaker is designed to amplify the sound, its surface contains many perforations. The GS point cloud represents this wire-like surface structure as multiple disjoint slender Gaussian ellipsoids, with their main axes distributed along the boundaries between neighboring holes. In contrast, the Can, designed to hold goods, has a flat surface without any perforations. This flat surface is represented by several intersecting flat ellipsoids in the GS point cloud. By comparing the Gaussian ellipsoids, it becomes apparent that the structures, originally similar in the point cloud, have become significantly distinct. As a result, the right classification probabilities for both the Loudspeaker and the Can are both improved when using the GS point cloud.

The improvement brought by the GS point cloud not only exists in the classification of the Loudspeaker and the Can but also appears in classifying other objects with wire-like and flat surfaces such as the Bowl and the Basket (\textit{i.e.}, the yellow cycles in~\cref{fig:prob_evo}(a) and~\cref{fig:prob_evo}(c), and the second row in~\cref{fig:size_orientation_ellipsoid}), as well as the Loudspeaker and the Paper box (\textit{i.e.}, the third row in~\cref{fig:size_orientation_ellipsoid}). These improvements validate the theoretical analysis presented in~\cref{sec:amb_theory_analysis}.

\subsection{Features Visualization}
To better understand the effects of different components in GS point cloud-based 3D classification, we focus on the high-dimensional global feature in the point-wise MLP methods (please refer to~Fig.~2 of~\cite{qi2017pointnet,qi2017pointnet++} for details). We feed the entire dataset into PointNet and visualize the global feature using the t-SNE (t-Distributed Stochastic Neighbor Embedding) technique~\cite{JMLR:v9:vandermaaten08a}.

\cref{fig:tsne} displays our visualization results, demonstrating the improvement in clustering visualization as we varied the input from only position to include opacity, scale, and rotation, and finally incorporating all Gaussian coefficients. It is evident that with the inclusion of Gaussian-related coefficients, the overlap between different categories in the clustering graph decreases, boundaries become clearer, and instances within the same category are grouped more closely. Using only position information results in greater overlap, especially when dealing with similar shapes. However, when geometric features combined with other Gaussian coefficients are used as input, this overlap is significantly reduced, and different categories are distinctly separated. This phenomenon not only exists in the global feature produced by PointNet, but also appears in other classifiers. For further details, please refer to the supplementary material.

\section{Conclusion}
In this work, we have proposed the GS point cloud-based 3D classification. We have identified two inherent ambiguities in traditional point cloud-based 3D classification, \textit{i.e.}, the characterization of local shape and appearance. These ambiguities are independent of the subsequent classification models and thus cannot be effectively addressed. The GS point cloud represents 3D objects as Gaussian ellipsoids with transparency, where the scale, rotation, and opacity of the ellipsoids better characterize the object's local shapes and appearance, significantly mitigating these ambiguities. We have demonstrated the enhancement of the GS point cloud and its improvements in object classification, addressing the aforementioned ambiguities, using the first GS point cloud dataset we constructed.

Unlike traditional point cloud, which can be captured with a variety of devices, the acquisition of GS point cloud relies on multiple RGB cameras, thereby limiting its applicability to a broader range of 3D tasks. Consequently, it is necessary to develop new devices and algorithms for obtaining GS point cloud in the future.
{
    \small
    \bibliographystyle{ieeenat_fullname}
    \bibliography{main}
}

\end{document}